\documentclass{article}

\usepackage{arxiv}

\usepackage[utf8]{inputenc} 
\usepackage[T1]{fontenc}    
\usepackage{hyperref}       
\usepackage{url}            
\usepackage{booktabs}       
\usepackage{amsfonts}       
\usepackage{nicefrac}       
\usepackage{microtype}      
\usepackage{lipsum}
\usepackage{graphicx}
\graphicspath{ {./images/} }

\title{PatiGonit22K: A Comprehensive Dataset for Solving Complex Bengali MWPs}

\author{
 Swastika Kundu \\
  Department of Computer Science and Engineering\\
  Ahsanullah University of Science and Technology, Dhaka, Bangladesh\\
  \texttt{swastikakundu123@gmail.com} \\
   \And
 Azizul Hakim Fayaz \\
  Department of Computer Science and Engineering\\
  Southeast University, Dhaka, Bangladesh\\
  \texttt{azizulhakimfayaz@gmail.com} \\
  \And
 Tashreef Muhammad \\
  Department of Computer Science and Engineering\\
  Southeast University, Dhaka, Bangladesh\\
  \texttt{tashreef.muhammad@seu.edu.bd} \\
}

\begin{document}
\maketitle
\begin{abstract}
Mathematical Word Problems (MWPs) are an important benchmark for evaluating natural language understanding and quantitative reasoning. Despite recent progress in high resource languages, Bengali remains underexplored due to the limited availability of large scale annotated datasets. In this work, we introduce PatiGonit22K, an expanded Bengali MWP dataset containing 22,441 problems, developed by extending the original PatiGonit dataset with a substantially larger collection of complex mathematical problems. The dataset includes both simple and multi operation equations, providing a balanced benchmark for evaluating mathematical reasoning across different difficulty levels. Each problem is carefully translated, annotated, culturally adapted, and verified to ensure linguistic consistency and mathematical correctness. By increasing both the scale and complexity of Bengali MWPs, PatiGonit22K provides a more comprehensive resource for future research on mathematical reasoning and educational NLP applications in low resource languages
\end{abstract}


\section{Introduction}

Despite significant progress in mathematical word problem (MWP) research, most publicly available datasets and benchmark systems have been developed for high-resource languages such as English and Chinese. Bengali is spoken by more than 230 million people worldwide~\cite{paul2025geospatial} and is the seventh most spoken language globally, yet it remains a low-resource language for MWP research because of the limited availability of large-scale annotated datasets and domain-specific resources.

The introduction of the PatiGonit dataset marked the first benchmark for Bengali MWPs by providing 10,000 translated mathematical word problems. However, the dataset contains relatively few complex multi-operation problems, limiting its effectiveness in evaluating advanced mathematical reasoning. To address this limitation, we introduce \textbf{PatiGonit22K}, an expanded dataset containing 22,441 Bengali mathematical word problems with a substantially larger proportion of complex problems. The dataset is carefully translated, annotated, and verified to provide a more comprehensive benchmark for future Bengali MWP research.
\begin{figure}[t]
    \centering
    \includegraphics[width=0.48\textwidth]{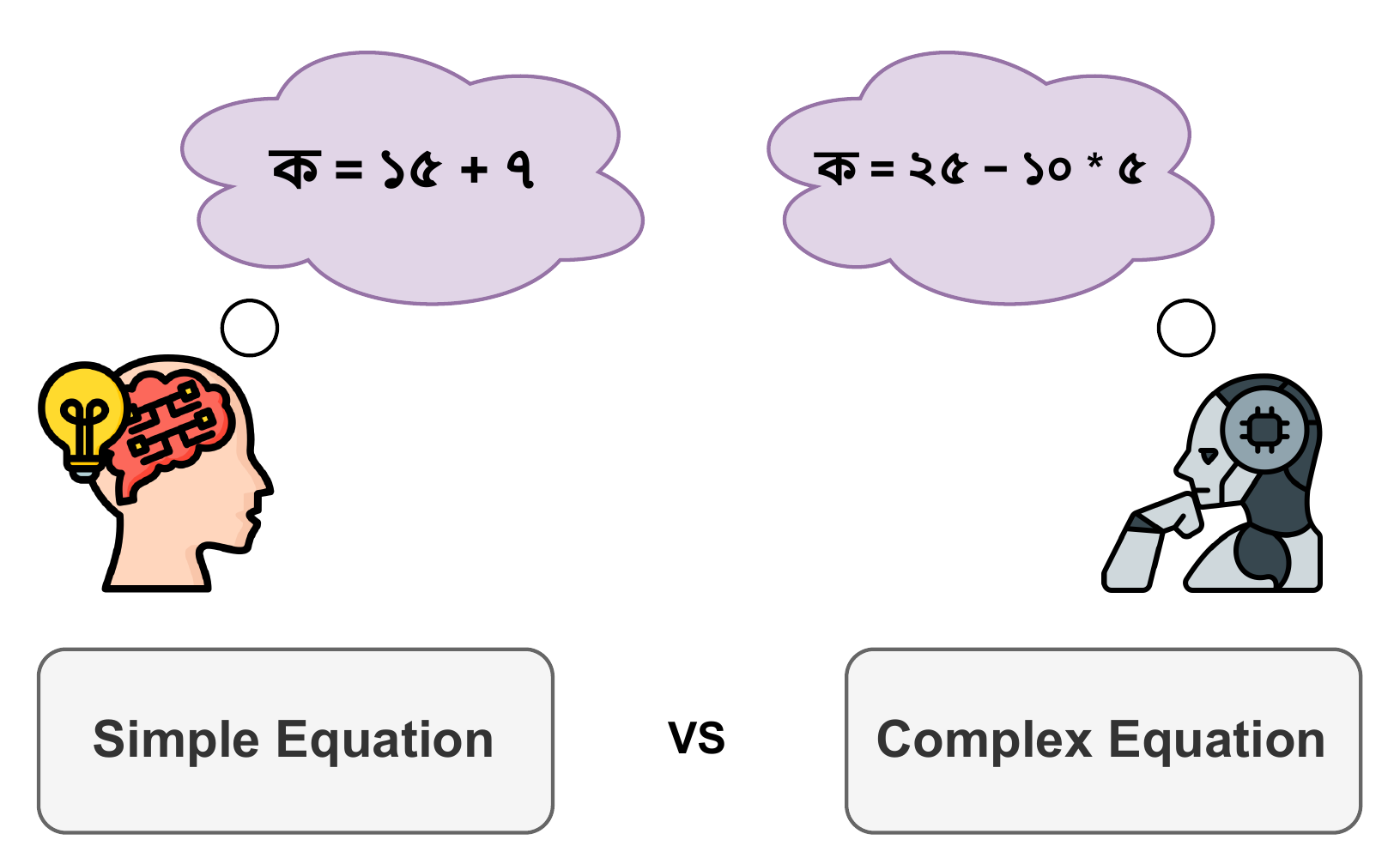}
    \caption{An example of Simple and Complex Equation}
    \label{fig:intro}
\end{figure}

\section{Related Work}

The introduction of the \textit{PatiGonit} dataset marked the first benchmark for Bengali Mathematical Word Problems (MWPs), containing 10,000 translated arithmetic problems adapted from the MAWPS dataset . The authors evaluated several transformer-based models, including Basic Transformer, mT5, BanglaT5, and mBART50, for generating mathematical equations from Bengali text. Among them, mT5 achieved the highest accuracy of 97.30\%, demonstrating the effectiveness of transformer architectures for Bengali MWP solving and establishing a strong baseline for future research\cite{era2024empowering}

To improve reasoning over complex Bengali MWPs, Paul \textit{et al.} introduced the \textit{SOMADHAN} dataset, consisting of 8,792 human-annotated mathematical word problems with step-by-step solutions \cite{paul2025leveraging}. The study investigated several large language models, including GPT-4o, GPT-3.5 Turbo, Llama, DeepSeek, and Qwen, using Chain-of-Thought (CoT) prompting and LoRA fine-tuning. Their experiments showed that CoT reasoning consistently improved mathematical reasoning performance, with Llama-3.3 70B achieving the highest accuracy of 88\% under few-shot settings.

The \textit{GSM-Plus-BN} benchmark was proposed as the first perturbation-based Bangla mathematical reasoning dataset for evaluating the robustness of large language models \cite{paul2026gsm}. The benchmark evaluates models including Qwen3, Llama, and GPT-OSS using both standard prompting and Chain-of-Thought prompting. Experimental results demonstrated that GPT-OSS-20B achieved the highest overall accuracy of 96.08\%, while Llama-3.3-70B and GPT-OSS-120B exhibited stronger robustness against linguistic perturbations.

Mondal \textit{et al.} introduced \textit{BMWP}, the first Bengali mathematical word problem dataset designed for arithmetic operation prediction and problem solving, containing 8,653 annotated problems \cite{mondal2025bmwp}. The study compared several deep learning architectures, including LSTM, Bi-LSTM, GRU, BGRU, SimpleRNN, Stacked LSTM, and GPT-2. Among these models, LSTM achieved the best validation accuracy of 90.42\%, demonstrating the effectiveness of recurrent neural networks for Bengali MWP solving.

Aurpa \textit{et al.} proposed \textit{Shomikoron}, the first Bangla dataset for mathematical equation recognition, comprising 3,430 mathematical statements paired with their corresponding equations \cite{aurpa2024shomikoron}. The authors evaluated several transformer-based models, including mBERT, ELECTRA, XLNet, RoBERTa, and DistilBERT, for equation prediction. Experimental results showed that mBERT achieved the best performance with an accuracy of 99.8\%, highlighting the effectiveness of transformer models for extracting mathematical equations from Bangla text.

\section{Dataset}

\subsection{Data Collection}

To address the scarcity of Bengali Math Word Problem (MWP) datasets, we expanded the existing PatiGonit dataset from 10,000 to 22,441 problems. The additional problems were curated to increase the proportion of complex and overall number of samples, thereby enhancing the dataset’s diversity and representativeness. The dataset was collected from multiple sources to ensure diversity in problem types and difficulty levels. English MWPs from the \textbf{MAWPS Dataset (mawps-single)}~\cite{koncel2016mawps} were translated into Bengali with careful attention to linguistic fidelity and mathematical correctness. Arithmetic problems were also adapted from the \textbf{MATH23K Dataset}~\cite{wu2021math}, ensuring accurate representation of numerical values and operators in Bengali. Additionally, single-operation problems from the \textbf{MAWPS Dataset}~\cite{koncel2016mawps} were incorporated to increase coverage of simpler problem types. From these sources, both \textit{simple} (single-operation) and \textit{complex} (multi-operation) problems were collected, creating a dataset that is well-suited for evaluating mathematical reasoning across varying levels of difficulty.

\subsection{Dataset Statistics}

The expanded dataset, referred to as \textbf{PatiGonit22K}, contains a total of 22,441 problems, including both \textit{simple} and \textit{complex} equations. The distribution of problems based on equation type is summarized in Table~\ref{tab:dataset_stats}.

\begin{table}[h!]
\centering
\caption{Statistics of the PatiGonit22k Dataset}
\label{tab:dataset_stats}
\begin{tabular}{|l|c|c|}
\hline
\textbf{Equation Type} & \textbf{Count} & \textbf{Examples of Operations} \\
\hline
Simple Addition & 1419 & + \\
Simple Subtraction & 1266 & - \\
Simple Multiplication & 1303 & * \\
Simple Division & 1391 & / \\
Complex (Mixed) & 17,029 & +, -, *, / combined \\
\hline
\textbf{Total Simple Equations} & 5412 & - \\
\textbf{Total Equations Processed} & 22,441 & - \\
\hline
\end{tabular}
\end{table}

Out of the 22,441 problems, 5,412 are simple equations involving a single operation (addition, subtraction, multiplication, or division), while 17,029 are complex equations that combine multiple operations. This distribution ensures sufficient coverage of both basic and advanced mathematical reasoning tasks, making \textbf{PatiGonit22K} a robust benchmark for Bengali MWPs.

\subsection{Dataset Annotation}

Each problem in \textbf{PatiGonit22k} was carefully translated to ensure high-quality training and evaluation. The annotation process followed these rules:

\begin{enumerate}
    \item \textbf{Translation and Cultural Adaptation:} Problems, including both the question text and the corresponding equations, were translated from English into Bengali while maintaining linguistic fidelity. Cultural relevance was preserved by adapting numerals, units, and symbols appropriately for Bengali-speaking students.
    
    \item \textbf{Answer Column:} A separate column was added to the dataset containing the solution to each equation, enabling straightforward evaluation of model predictions.
    
    
    \item \textbf{Quality Verification and Consistency Checks:} All annotations were verified by a team of bilingual annotators proficient in both Bengali and English mathematical terminology. The dataset was checked for duplicate problems, inconsistent equation representations, and ambiguous wording. Problems failing these checks were corrected or removed.
\end{enumerate}

\begin{figure}[t]
\centering
\includegraphics[width=1.0\columnwidth]{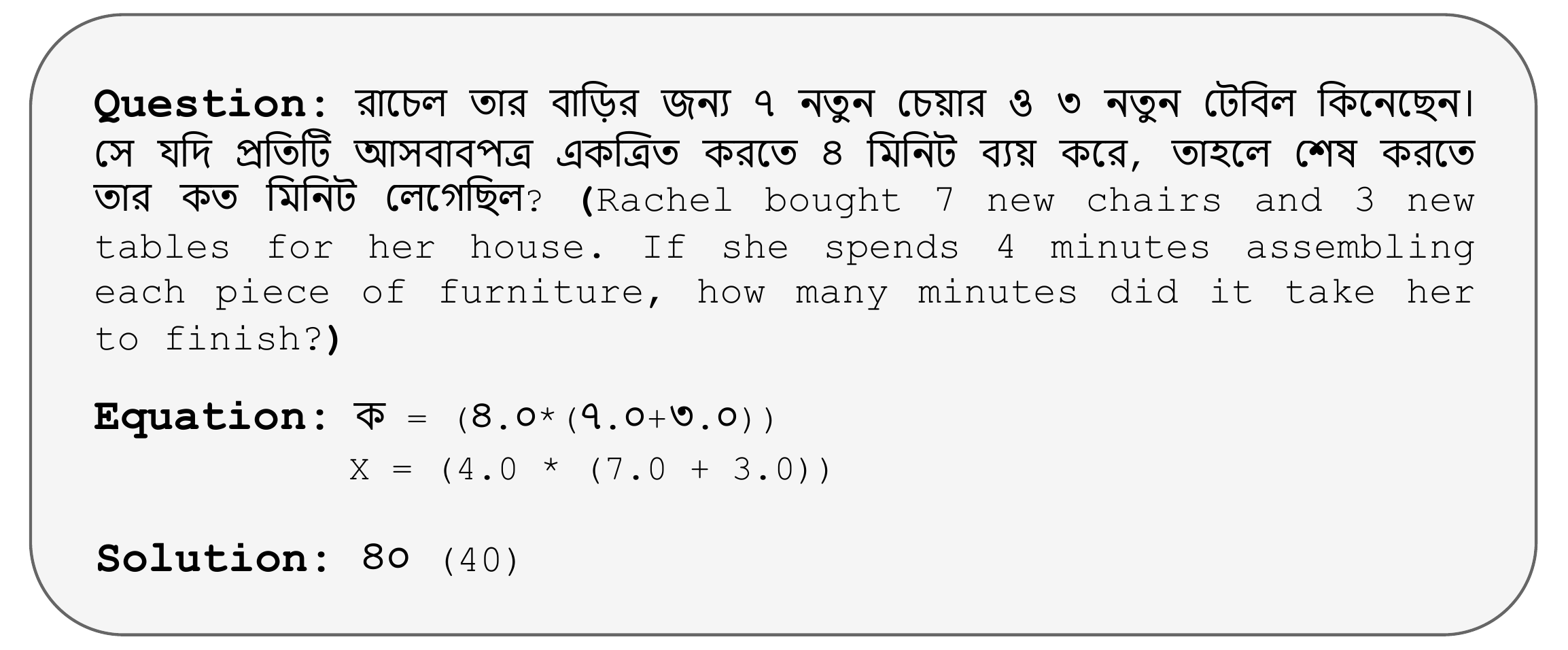} 
\caption{An example of PatiGonit22k dataset}
\label{fig:dataset}
\end{figure}

This rigorous annotation ensures that \textbf{PatiGonit22k} provides a reliable benchmark for evaluating transformer-based models on both simple and complex Bengali MWPs, supporting robust research in low-resource mathematical NLP tasks. Figure~\ref{fig:dataset} illutrates an overview of our proposed dataset.

\section*{Declarations}


\begin{itemize}
\item Data availability: The PatiGonit22K dataset, which contains complex Bengali grade school math word problems, is publicly available for research and academic purposes. Researchers interested in utilizing the dataset can access it through the following link: \href{https://data.mendeley.com/datasets/pgxv75scs7/1}{Dataset of PatiGonit22K (Original Data) (Mendeley Data)}.
\end{itemize}

\bibliographystyle{unsrt}  

\bibliography{references}

@inproceedings{era2024empowering,
  title={Empowering bengali education with ai: Solving bengali math word problems through transformer models},
  author={Era, Jalisha Jashim and Paul, Bidyarthi and Aothoi, Tahmid Sattar and Zim, Mirazur Rahman and Shah, Faisal Muhammad},
  booktitle={2024 27th International Conference on Computer and Information Technology (ICCIT)},
  pages={909--914},
  year={2024},
  organization={IEEE}
}

@article{paul2025leveraging,
  title={Leveraging large language models for bengali math word problem solving with chain of thought reasoning},
  author={Paul, Bidyarthi and Era, Jalisha Jashim and Zim, Mirazur Rahman and Aothoi, Tahmid Sattar and Shah, Faisal Muhammad},
  journal={arXiv preprint arXiv:2505.21354},
  year={2025}
}

@article{paul2025geospatial,
  title={Geospatial and Temporal Trends in Urban Transportation: A Study of NYC Taxis and Pathao Food Deliveries},
  author={Paul, Bidyarthi and Chowdhury, Fariha Tasnim and Biswas, Dipta and Sultana, Meherin},
  journal={arXiv preprint arXiv:2505.03816},
  year={2025}
}

@article{paul2026gsm,
  title={GSM-Plus-BN: A Perturbation-Based Benchmark for Bangla Mathematical Reasoning in Large Language Models},
  author={Paul, Bidyarthi and Mayouree, Nahida Jannat and Karim, Md Asif and Nath, Sagar Chandra and Kundu, Swastika},
  journal={arXiv preprint arXiv:2607.13248},
  year={2026}
}

@article{mondal2025bmwp,
  title={BMWP: the first Bengali math word problems dataset for operation prediction and solving},
  author={Mondal, Sanchita and Khatua, Debnarayan and Mandal, Sourav and Prasad, Dilip K and Sekh, Arif Ahmed},
  journal={Discover Artificial Intelligence},
  volume={5},
  number={1},
  pages={25},
  year={2025},
  publisher={Springer}
}

@article{aurpa2024shomikoron,
  title={Shomikoron: dataset to discover equations from bangla mathematical text},
  author={Aurpa, Tanjim Taharat and Fariha, Kazi Noshin and Hossain, Kawser},
  journal={Data in Brief},
  volume={55},
  pages={110742},
  year={2024},
  publisher={Elsevier}
}

@inproceedings{koncel2016mawps,
  title={MAWPS: A math word problem repository},
  author={Koncel-Kedziorski, Rik and Roy, Subhro and Amini, Aida and Kushman, Nate and Hajishirzi, Hannaneh},
  booktitle={Proceedings of the 2016 conference of the north american chapter of the association for computational linguistics: human language technologies},
  pages={1152--1157},
  year={2016}
}

@inproceedings{wu2021math,
  title={Math word problem solving with explicit numerical values},
  author={Wu, Qinzhuo and Zhang, Qi and Wei, Zhongyu and Huang, Xuan-Jing},
  booktitle={Proceedings of the 59th Annual Meeting of the Association for Computational Linguistics and the 11th International Joint Conference on Natural Language Processing (Volume 1: Long Papers)},
  pages={5859--5869},
  year={2021}
}

\end{document}